\PassOptionsToPackage{table}{xcolor}
\documentclass[  sigconf ]{acmart}

\makeatletter
\def\@copyrightspace{\relax}
\makeatother
\setcopyright{none}

\usepackage{graphicx}

\usepackage{booktabs}
\usepackage{rotating}
\usepackage{xfrac}

\usepackage{hyperref}

\usepackage[acronym]{glossaries}
\makeglossaries
\newacronym{tbn}{TBN}{Tiled Bit Network}
\newacronym{dnn}{DNN}{Deep Neural Network}
\newacronym{bnn}{BNN}{Binary Neural Network}

\newacronym{lth}{LTH}{Lottery Ticket Hypothesis}
\newacronym{bnns}{BNN's}{Binary Neural Networks}

\newacronym{mpts}{MPT's}{Multi-Prize Tickets}
\newacronym{smc}{SMC}{Simple Matching Coefficient}

\newacronym{ji}{JI}{Jaccard Index}
\newacronym{sgd}{SGD}{Stochastic Gradient Descent}
\newacronym{imp}{IMP}{Iterative Magnitude Pruning}
\newacronym{iwr}{IWR}{Iterative Weight Recycling}

\newacronym{iot}{IoT}{Internet of Things}
\newacronym{ai}{AI}{Artificial Intelligence}
\newacronym{nlp}{NLP}{natural language processing}

\newacronym{dl}{DL}{Deep Learning}

\newacronym{ml}{ML}{machine learning}

\newacronym{qnns}{BNN's}{Quantized Neural Networks}

\newacronym{sbt}{SBT}{Sparse Binary Transformer}
\newacronym{sbnn}{SBNN}{Sparse Binary Neural Networks}
\newacronym{pot}{POT}{Peak Over Threshold}

\newacronym{flops}{FLOPs}{floating point operations}

\newacronym{lstf}{LSTF}{long-term time series forecasting}

\newacronym{fp32}{FP32}{floating point 32}

\newacronym{gpu}{GPU}{Graphics Processing Unit}

\newacronym{qat}{QAT}{Quantization Aware Training}

\newacronym{cnn}{CNN}{Convolutional Neural Network}

\newacronym{fsmn}{FSMN}{Feedforward Sequential Memory Network}
\newacronym{fps}{FPS}{Frames Per Second}

\newacronym{iou}{IoU}{Intersection Over Union}

\newacronym{bwnn}{BWNN}{Binary Weight Neural Network}

\newacronym{vit}{ViT}{Vision Transformer}

\usepackage{epsfig}
\usepackage{comment}
\usepackage{graphicx}

\usepackage{booktabs}
\usepackage{makecell}
\usepackage{multirow}
\usepackage{bm}
\usepackage{arydshln}

\copyrightyear{}
\acmYear{}
\acmDOI{}

\acmConference[CIKM '24]{}{October 21--25,
  2024}{Boise, ID}
\acmISBN{}

\usepackage[capitalize,noabbrev]{cleveref}

\theoremstyle{plain}

\theoremstyle{definition}

\theoremstyle{remark}

\usepackage[textsize=tiny]{todonotes}

\usepackage{makecell, cellspace, caption}

\usepackage{cite}
\usepackage{amsmath,amsfonts}
\usepackage{algorithmicx}
\usepackage{graphicx}
\usepackage{textcomp}

\definecolor{aliceblue}{rgb}{0.94, 0.97, 1.0}
\definecolor{beaublue}{rgb}{0.74, 0.83, 0.9}

\AtBeginDocument{%
  \providecommand\BibTeX{{%
    \normalfont B\kern-0.5em{\scshape i\kern-0.25em b}\kern-0.8em\TeX}}}

\usepackage{listings}

\usepackage{graphicx}

\usepackage{booktabs}
\usepackage{rotating}
\usepackage{xfrac}

\usepackage{hyperref}

\usepackage{times}
\usepackage{epsfig}

\usepackage{graphicx}

\usepackage{booktabs}
\usepackage{makecell}
\usepackage{multirow}
\usepackage{bm}
\usepackage{arydshln}

\usepackage[utf8]{inputenc} 
\usepackage[T1]{fontenc}    
\usepackage{hyperref}       
\usepackage{url}            
\usepackage{booktabs}       
\usepackage{nicefrac}       
\usepackage{microtype}      

\usepackage{wrapfig}
\usepackage{booktabs}
\usepackage{tabularx}
\usepackage{makecell}

\usepackage{caption}
\usepackage{subcaption}
\usepackage{caption,stackengine}
\usepackage{enumitem}

\usepackage{capt-of}
\usepackage{booktabs}
\usepackage{varwidth}
\usepackage{framed}

\usepackage{xpatch}

\usepackage{microtype}
\usepackage{graphicx}
\usepackage{booktabs} 

\usepackage{hyperref}

\usepackage{algorithm}
\usepackage{algpseudocode}

\usepackage{etoolbox}

\usepackage{mathtools}

\newcommand{\ffrac}[2]{\ensuremath{\frac{\displaystyle #1}{\displaystyle #2}}}

\begin{document}

\title{Tiled Bit Networks: Sub-Bit Neural Network Compression Through Reuse of Learnable Binary Vectors}

\author{Matt Gorbett}
\email{matt.gorbett@colostate.edu}
\affiliation{%
  \institution{Colorado State University}
  \city{Fort Collins, CO}
  \country{USA}
}

\author{Hossein Shirazi}
\email{hshirazi@sdsu.edu}
\affiliation{%
  \institution{San Diego State University}
  \city{San Diego, CA}
  \country{USA}
}

\author{Indrakshi Ray}
\email{indrakshi.ray@colostate.edu}
\affiliation{%
  \institution{Colorado State University}
  \city{Fort Collins, CO}
  \country{USA}
}


\begin{abstract}
\glspl{bnn} enable efficient deep learning by saving on storage and computational costs.  
However, as the size of neural networks continues to grow, meeting computational requirements remains a challenge. 
In this work, we propose a new form of quantization to tile neural network layers with sequences of bits to achieve sub-bit compression of binary-weighted neural networks.  The method learns binary vectors (i.e. tiles) to populate each layer of a model via aggregation and reshaping operations. During inference, the method reuses a single tile per layer to represent the full tensor. We employ the approach to both fully-connected and convolutional layers, which make up the breadth of space in most neural architectures.  Empirically, the approach achieves near full-precision performance on a diverse range of architectures (CNNs, Transformers, MLPs) and tasks (classification, segmentation, and time series forecasting) with up to an 8x reduction in size compared to binary-weighted models. We provide two implementations for Tiled Bit Networks: 1) we deploy the model to a microcontroller to assess its feasibility in resource-constrained environments, and 2) a GPU-compatible inference kernel to facilitate the reuse of a single tile per layer in memory. 

\end{abstract}








\maketitle

\glsreset{bnn}

\section{Introduction}

The progress of modern machine learning can be largely attributed to the exponential growth of \glspl{dnn}. Empirically, the capacity of \glspl{dnn} is expanding at an astounding rate \citep{brown2020language}, a practice supported by 
theory showing that sufficiently over-parameterized models are in fact necessary for deep learning \citep{hu2021model,allen2019learning}.
Alongside this progress, the growing presence of resource-constrained machines (e.g. embedded devices, cell phones) has created unique opportunities to deploy increasingly large \glspl{dnn} in novel environments. Consequently, maximizing the computational efficiency of neural networks is a relevant challenge at various scales of application. 

Efforts toward efficient deep learning span a broad range of techniques such as architectural design \citep{howard2017mobilenets, sandler2018mobilenetv2}, neural architecture search \citep{lin2020mcunet}, knowledge distillation \citep{hinton2015distilling, sanh2019distilbert}, and quantization \citep{hubara2017quantized, zhou2017incremental, choukroun2019low}.  Quantization, which converts high precision neural network weights into discrete values, has achieved success in practical applications \citep{nagel2021white, rusci2020memory}, and has been applied down to the scale of \glspl{bnn} where the weights (and often activations) of a model are single bit values \citep{courbariaux2016binarized}.

\begin{figure}[t]
\centering
\includegraphics[width=8cm]{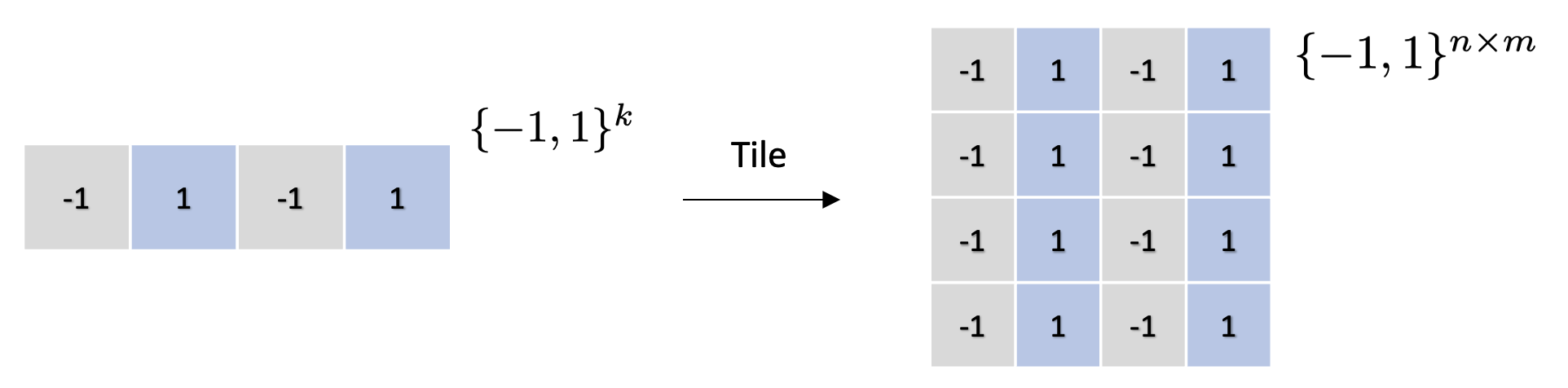}
\caption{\textbf{Tiling Illustration:} A binary tile (left) of size $k=4$ is replicated four time  to create a weight matrix of size 16 (right). Tiling is used during the training process of \glspl{tbn} to learn vectors for populating the parameters of a model (as illustrated above). During inference, only a single tile needs to be referenced per layer (left) -- a specialized kernel can reuse the tile throughout layer computation for memory savings.}\label{tiling_teaser}
\end{figure}

While \glspl{bnn} have been established as a practical and extreme form of quantization, an emerging line of research has gone a step further with \textit{sub-bit} neural network compression, which requires less than a single bit per model parameter. Wang et. al \citep{wang2021sub} first observed that the discrete set of binary convolutional kernels tend to cluster into a smaller subset; as a result, they devised a training regime to use a smaller set of kernels. Subsequent work has achieved improved compression by leveraging properties of binary convolutional kernels using minimum spanning trees \citep{vo2023mst} and sparse kernel selection \citep{wang2023compacting}. 

\begin{figure*}[t]
\centering
\includegraphics[width=17cm]{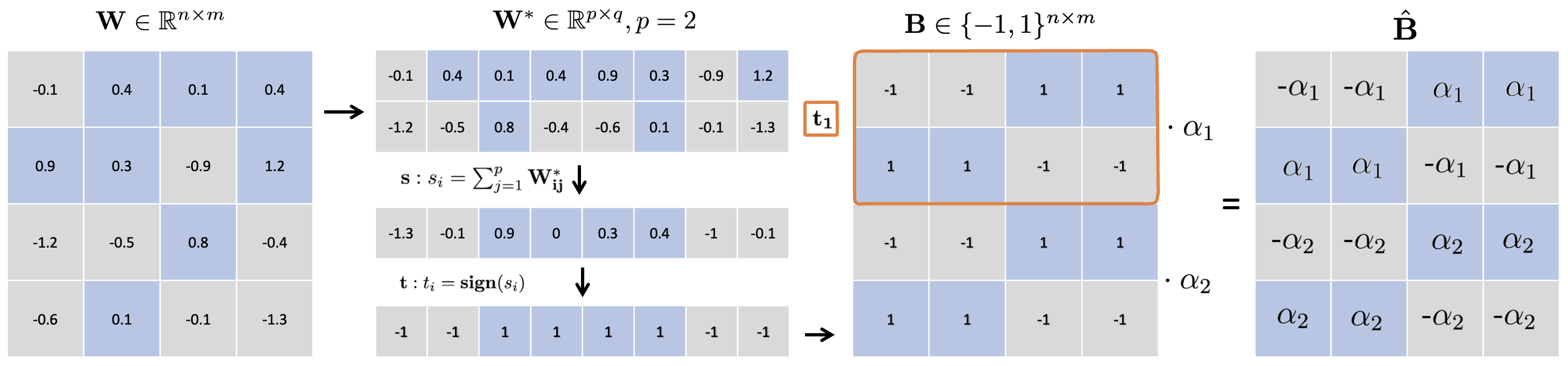}
\caption{\textbf{Tile Construction During Training:} For each layer of a neural network, we train a standard weight tensor ($\mathbf{W}$) (left).
During the training, we compress the parameter by a factor of $p$ by performing a reshaping (second column top) and then sum operation (second column middle). We use the straight-through estimator to binarize the vector $\mathbf{s} $, creating the tile $\mathbf{t}$ (bottom of the second column). We next create binary weights $\mathbf{B}$ from the resulting binary vector by tiling vector $\mathbf{t}$ two times and reshaping it to an $n \times m$ tensor (third column). Finally, we apply a scalar $\alpha$ over each of the two tiles, resulting in the final weight tensor $\mathbf{\hat{B}}$. During the inference, only a single tile is needed, along with a small number of $\alpha$ scalars.  }\label{scoring}
\end{figure*}

Independent from previous approaches, this work proposes tiling neural networks with binary weights to achieve sub-bit memory and storage compression of model parameters. \glsdisp{tbn}{Tiled Bit Networks (TBNs)} learn binary sequences (tiles) to fill in the weights of a \glspl{dnn} layers during training. A tiling operation is depicted in Figure \ref{tiling_teaser}. The algorithm learns a condensed  parameter representation for each layer by compressing the weight values using tensor reshaping and aggregation (Figure \ref{scoring}); a scalar is additionally applied to each layer or tile. 

\begin{figure}[t]
\centering
\includegraphics[width=8cm]{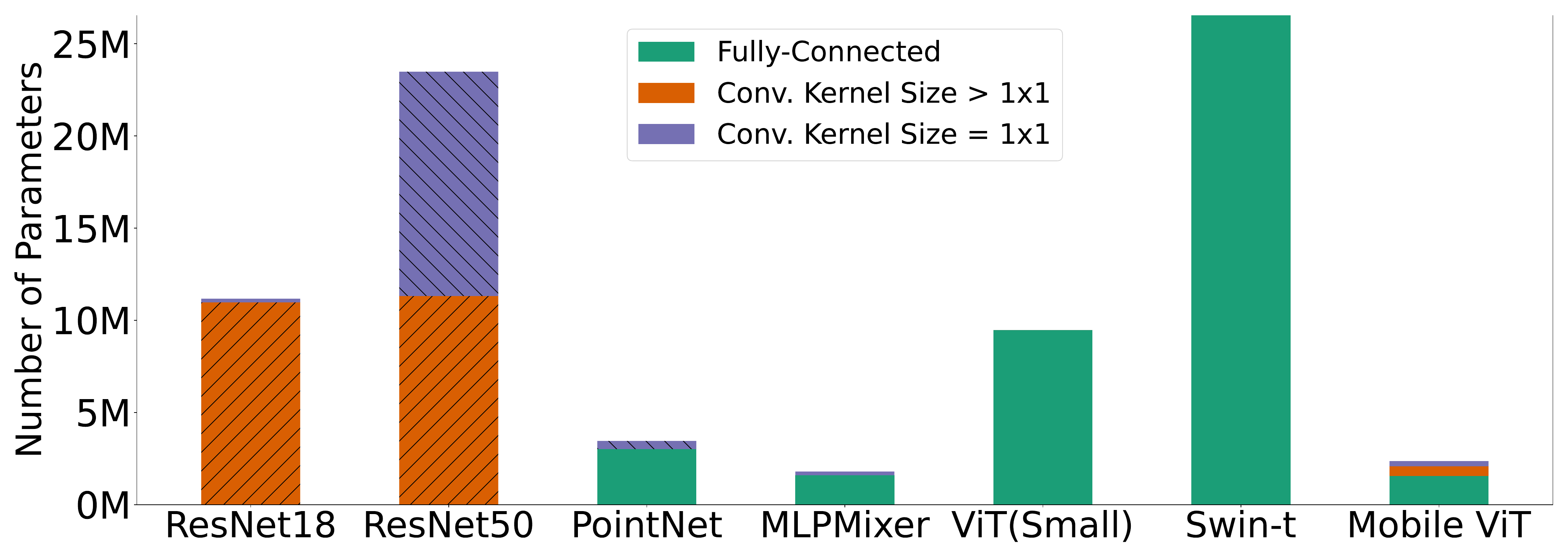}
\caption{ 
\textbf{Composition of popular \glspl{dnn}:} The ResNet series is made up primarily of convolutional layers; MLPs (PointNet, MLPMixer) and Transformers (Swin-t, ViT, Mobile ViT) consist mostly of fully-connected parameters.  
}\label{barchart}
\end{figure}

Unique from previous work that leverages the properties of convolutional kernels to achieve sub-bit compression, \glspl{tbn} work on both fully-connected and convolutional layers, a relevant application for modern architectures as depicted in Figure \ref{barchart}. We test the approach on CNNs, Transformers, PointNets, and MLPs, enabling for the first time sub-bit \gls{dnn} compression on models with high proportions of fully-connected parameters. Compared to previous approaches, \glspl{tbn} achieve better or similar performance on image classification tasks. Empirically, across 2D and 3D image classification, \glspl{tbn} achieve performance on par with \glspl{bwnn} with only a fraction of the total parameters -- less than a single bit is stored and accessed per model parameter. \glspl{tbn} additionally achieve strong performance on semantic and part segmentation tasks as well as time series forecasting. 

We provide two implementations for model inference in Section \ref{imp}, which both require only a single tile per model layer in memory.  First, we implement a lightweight \gls{tbn} for deployment on a microcontroller, showing that the algorithm reduces memory and storage consumption compared to \glspl{bwnn}. We also implement a GPU-compatible inference kernel using the Triton library \citep{triton_tillet}, which allows memory savings through the reuse of a single tile in the highly parallelized setting. The \gls{tbn} inference kernel requires 2.8x less peak memory (78.5MB vs. 222.5MB) compared to a standard kernel on ImageNet \gls{vit} (Small) when both have full precision weights.





Our contributions are as follows: 
\vspace{-.5em}
\begin{itemize}
    \item We achieve sub-bit memory and storage compression of neural network parameters by learning sequences of binary values (tiles) to populate the layers of \glspl{dnn}. We apply the method to fully-connected and convolutional layers. To the best of our knowledge, this is the first work to show substantial sub-bit compression of fully-connected \glspl{dnn} which are relevant in Transformers and MLPs (PointNet, MLPMixer). 
  
    \item We provide two implementations that achieve sub-bit compression of model parameters by \textit{reusing} a single tile per model layer: 1) we deploy a \gls{tbn} to a microcontroller with a customized C kernel, and 2) we develop a specialized GPU kernel for fully-connected layers to leverage memory savings of tiled parameters during inference.   
\end{itemize}

\label{intro}

\section{Related Work}

\textbf{Quantized and Binary Neural Networks }
\gls{dnn} quantization reduces full-precision weights and activations to discrete and lower precision values to enhance model storage, memory, and inference speed \citep{zhuang2018towards, lin2016fixed}. The most extreme quantization
was conventionally thought to be binarization, where weights can only be $\pm1$ \citep{qin2020binary}. Binarization helps reduce computation, however, it often reduces model accuracy. Several works attempt to alleviate this issue such as XNOR-Net, which used channel-wise scaling factors
for \glspl{bwnn} and \glspl{bnn} \citep{rastegari2016xnor}. IR-Net \citep{qin2020forward} preserved the information by maximizing the entropy of the information while minimizing the quantization error. ReActNet used generalized activation functions to get within 3\% of full-precision accuracy on ImageNet \citep{liu2020reactnet}; Shang et al. utilized contrastive learning to learn \glspl{bnn} \citep{shang2022network}. Xu et al. proposed FDA, which estimates sign function gradients in a Fourier frequency domain \citep{xu2021learning}; Xu et al. proposed ReCU which introduces a rectified clamp unit to address dead weights \citep{9710916}.

We note that the \gls{bnn} research covered in this section uses binary activations as well as binary weights, and as a result, achieves significant memory and speed improvements. \glspl{tbn} use full-precision activations, however, still achieve storage and memory improvements from using a single tile per layer. We denote \glspl{bnn} that have full-precision activations as \glsdisp{bwnn}{Binary Weight Neural Networks (BWNNs)}. We indicate whether previous \gls{bnn} algorithms have binary activation's in benchmark experiments in Section \ref{experiments}.

\textbf{Sub-Bit Quantization}
Sub-bit \gls{dnn} compression reduces model sizes to less than a single bit per model parameter. Kim et al. \citep{8060432} proposed a kernel decomposition to reduce computations in binary CNNs. FleXOR \citep{lee2020flexor} used an encryption technique to store binary sequences.  Wang et al. \citep{wang2021sub} observed that the full set of binary convolutional kernels tends to cluster into a subset; they formulate a training technique to find the best subsets of kernels. Lan et al. \citep{lan2021compressing} stack convolutional filters to achieve sub-bit compression. Wang et al. \citep{wang2023compacting} group kernels into binary codebooks for sparse kernel selection. Finally, Vo et al. \citep{vo2023mst} propose minimum spanning tree compression, which takes advantage of the observation that output channels in binary convolutions can be computed
using another output channel and XNOR operations. 

Previous sub-bit compression approaches are distinct from \glspl{tbn}: initial work was based on removing redundancy and encrypting weights; CNN-based approaches are based on utilizing  properties of binary convolutional kernels.  \glspl{tbn}, on the other hand, achieve substantial compression on both fully-connected and convolutional layers, and can be applied to multiple architectures (CNNs, Transformers, MLPs). 



 \textbf{Efficient Machine Learning}
 Model quantization is a sub-field of efficient deep learning, which encompasses multiple areas not covered in this work such as low rank factorization \citep{li2018learning, choromanski2020rethinking,lebedev2014lowrank}, structured and unstructured pruning \citep{he2017channel, frankle2018lottery, gorbett2023sparse, gorbett2023randomly}, knowledge distillation \citep{hinton2015distilling, romero2014fitnets}, and memory efficiency through input patching \citep{lin2021mcunetv2} and attention tiling \citep{dao2022flashattention}. 

    \textbf{Embedded and On-Device Machine Learning} 
The size and computational requirements of \glspl{dnn} has motivated researchers to improve the compatibility of large models with hardware such as mobile phones and embedded devices (e.g. FGPAs, IoT Sensors)
\citep{cheng2017survey}. Architectural optimizations such as MobileNet \citep{sandler2018mobilenetv2}, ShuffleNet \citep{ma2018shufflenet}, and MCUNet \citep{lin2020mcunet, lin2022device} have been achieved success, including to ease memory constraints via layer patching \citep{lin2021mcunetv2}. 

\section{Method}\label{methods}

\glsdisp{tbn}{Tiled Bit Networks} are constructed from a standard neural network with layers $1,2,l ...,L$. 
We consider fully-connected and convolutional layers in this work since these layers generally make up the breadth of \glspl{dnn} weights. We do not consider bias parameters in this work.  

In this section, we describe the training process for \glspl{tbn}, which involves learning full-precision parameters ($\mathbf{W}$) and applying aggregation and reshaping to create the tile vectors $\mathbf{t}$. We then describe our approach to tile-wise scaling, the second step of training \glspl{tbn}. Finally, we describe training hyperparameters and their default settings.

\textbf{Layer-Wise Tiling }
The key to our approach is that we learn tile vectors $\mathbf{t}^{[1]},\mathbf{t}^{[l]},...,\mathbf{t}^{[L]}$ for each layer of our network. We initialize our model with full-precision values for each layer similar to standard training, creating a weight tensor $\mathbf{W}^{[l]}\in\mathbb{R}^{d_1 \times...d_k}$ for layer $l$, where $d_k$ is the dimensionality of the tensor (e.g., a fully-connected layer has $k=2$).  
The total of elements in the tensor is $N=\prod_{i=1}^{k} d_i$. 
During training, we update $\mathbf{W^{[l]}}$ via stochastic gradient descent. Our goal is to  \textit{compress} $\mathbf{W^{[l]}}$ by a factor of $p$, where size $N$ is divisible by $p$ such that $p \times q=N$. 
To achieve this we reshape tensor $\mathbf{W^{[l]}}$ as a
$p \times q$ dimensional matrix $\mathbf{W^{[l]*}}$ during forward propagation:

 \begin{equation}\label{eq:eq1}
\mathbf{W^{[l]}}\in\mathbb{R}^{d_1,...d_k}\rightarrow \mathbf{W^{[l]*}}\in\mathbb{R}^{p \times q} 
 \end{equation}

We then sum the reshaped weight tensor $\mathbf{W^{[l]*}}$ along the $p$ dimension to create a vector $\mathbf{s}\in\mathbb{R}^q$:

 \begin{equation}\label{eq:eq2}
\mathbf{s} = \begin{bmatrix}
\sum_{j=1}^{p} \mathbf{W^{[l]*}_{1j}} \\
\sum_{j=1}^{p} \mathbf{W^{[l]*}_{2j}} \\
\vdots \\
\sum_{j=1}^{p} \mathbf{W^{[l]*}_{qj}}
\end{bmatrix} = 
\begin{bmatrix}
s_1 \\
s_2 \\
\vdots \\
s_q 
\end{bmatrix} 
 \end{equation}

We next create tile $\mathbf{t^{[l]}}=[t_1, t_2,t_i...t_q]$ for a given layer by applying a threshold function to determine the binary value for each $s_i$ in  $\mathbf{s}$:

\begin{equation}\label{eq:eq3}
    t_i = 
\begin{cases}
    1, & \text{if } s_i > 0 \\
   -1, & \text{otherwise}
\end{cases} 
\end{equation}

Tile $\mathbf{t^{[l]}}$ is then replicated $p$ times to create  tile vector $\mathbf{b^{[l]}}\in\mathbb{R}^{N}$. Formally, let $\mathbf{1_N}$ be a vector of ones with size $N$. The tiling operation creates vector $\mathbf{b^{[l]}}$ as: 
 \begin{equation}
  \mathbf{b^{[l]}}=\mathbf{1_N} \otimes \mathbf{t^{[l]}} \label{eq:eq4}
 \end{equation}

where $\otimes$ is the Kronecker product. We create our final binary weight tensor $\mathbf{B^{[l]}}\in\{-1,1\}^{d_1,...d_k}$ by reshaping vector $\mathbf{b^{[l]}}$:


 \begin{equation}
  \mathbf{B^{[l]}}=\mathbf{vec^{-1}_{d_1,...d_k}(\mathbf{b^{[l]}})} \label{eq:eq5}
 \end{equation} 
 
where $\mathbf{vec^{-1}_{d_1,...d_k}}(\cdot)$ denotes a vector to $k$-dimensional tensor operation.

 We note that computing binary parameters $\mathbf{B^{[l]}}$ involves non-differentiable operations during forward propagation. As a result, we utilize straight-through gradient estimation, where the gradients of the model are passed-through the non-differentiable operator during backpropagation \citep{bengio2013estimating}. To achieve this we implement \cref{eq:eq1,eq:eq2,eq:eq3,eq:eq4,eq:eq5} in the forward pass of a customized differentiation engine, and on backpropagation we pass the gradients through the customized module to update $\mathbf{W^{[l]}}$. 

 Putting it together, the tiled model $f(\cdot)$ can be trained with parameters $\mathbf{W^{[l]}}$ (to compute $\mathbf{B^{[l]}}$) and inputs $x$, producing an output $y$ which serves as a continuous, differentiable approximation of a tiled neural network. In the context of straight-through gradient estimation, \(y\) is used during backpropagation to compute the gradient of loss $\mathcal{L}$ with respect to the parameter $\mathbf{W^{[l]}}$:

 \begin{equation}
  \quad \quad \frac{\partial \mathcal{L}}{\partial \mathbf{W}^{[l]}}= \frac{\partial \mathcal{L}}{\partial 
  y^{[l]}} \cdot \frac{\partial y^{[l]}}{\partial \mathbf{W}^{[l]}}, \quad \frac{\partial y^{[l]}}{\partial \mathbf{W}^{[l]}}\approx\frac{\partial y^{[l]}}{\partial \mathbf{B}^{[l]}}
 \end{equation}

where $y^{[l]}$ is the output of layer $l$ prior to the activation.  $\frac{\partial y^{[l]}}{\partial \mathbf{B}^{[l]}}$ involves the thresholding, tiling, and reshaping operations.

\begin{figure}[t]
\centering
\includegraphics[width=6.8cm]{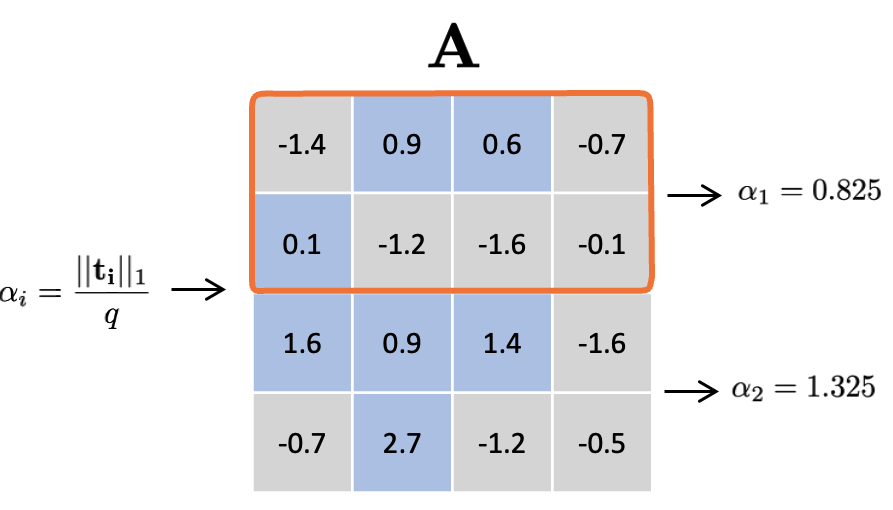}
\caption{We learn scalar $\alpha$  from tensor $\mathbf{A}$ by computing Equation \ref{eq6} or \ref{eq8} over its values ($\mathbf{W}$ can also be used in place of $\mathbf{A}$). The figure visualizes Equation \ref{eq8} which calculates $\alpha$ over each tile. }\label{alpha_construction}
\end{figure}

\textbf{Tile-wise Scalars} Similar to XNORNet \citep{rastegari2016xnor}, we scale $\mathbf{B}^{[l]}$ by $\alpha$. 
Rastegari et al. \citep{rastegari2016xnor} derived the optimal scaling factor of a binary weight filter as the average absolute value of a weight, a method widely used in other research \citep{martinez2020training,diffenderfer2020multi, qin2020binary}. We can use parameter $\mathbf{W^{[l]}}$ to compute the scalar since its non-aggregated size is the same as a standard weight tensor: 

\begin{equation}
    \alpha=\frac{||\mathbf{W^{[l]}}||_1}{N}\label{eq6}
\end{equation}

 We additionally experiment with an independent tensor, denoted $\mathbf{A^{[l]}}\in\mathbb{R}^{d_1 \times...d_k}$, to exclusively compute the $\alpha$ scalar.  We observe a slight performance benefit from using $\mathbf{A^{[l]}}$ in addition to $\mathbf{W^{[l]}}$. We add this option as a hyperparameter to our models.

Another hyperparameter setting for \glspl{tbn} calculates one $\alpha$ for each tile $\mathbf{t_{1}},\mathbf{t_2}...\mathbf{t_{p}}$ in layer $l$ by utilizing the $i^{th}$ set of the flattened tensor $\mathbf{A^{[l]}}$ (or $\mathbf{W^{[l]}}$), and calculating $\alpha_i$ using only these values. 
 This represents the optimal scalar for that particular tile. To do this, we can reshape 
 $\mathbf{A^{[l]}}$ to get the values corresponding to the $p^{th}$ tile of layer $l$:

\begin{equation}
   \mathbf{A^{[l]}}\in\mathbb{R}^{d_1,...d_k}\rightarrow \mathbf{A^{[l]*}}\in\mathbb{R}^{q \times p}\label{eq7}
\end{equation}

Next, similar to Equation \ref{eq6} we calculate the 1-norm for each segment of values corresponding to the $i^{th}$ tile in $\mathbf{A^{[l]}}$.  We divide this number by $q$ (the size of each tile) to give us $\alpha_1,\alpha_2,...\alpha_p$:

 \begin{equation}
\begin{bmatrix}
\alpha_1^{[l]} \\
\vdots \\
\alpha_p^{[l]}
\end{bmatrix} = \begin{bmatrix}
\ffrac{||\mathbf{A^{[l]*}_{1}}||_1}{q} \\
\vdots \\
\ffrac{||\mathbf{A^{[l]*}_{p}}||_1}{q} \\
\end{bmatrix} \label{eq8}
 \end{equation}

Each $\alpha$ gets multiplied element-wise by its corresponding tile in $\mathbf{B^{[l]}}$.  The resulting tensor, $\mathbf{\hat{B}^{[l]}}$ is used for the operation on the inputs. Equation \ref{eq8} is depicted in Figure  \ref{alpha_construction}. 

After training is complete, we save a vector of size $q$ for each layer along with full-precision scalars ($\alpha$s).  We describe our implementation in Section \ref{imp}.

\textbf{Hyperparameter Settings}
We test our models with several hyperparameter configurations to assure the best performance.  \glspl{tbn} primarily contains three hyperparameters: 

\begin{enumerate}
    \item Minimum layer size for tiling, $\lambda$.  We set a minimum size $N$ of a \gls{dnn} layer required for tiling to be performed. Tiling smaller layers causes a drop in performance (Figure \ref{ablation2}). 
    \textbf{Default:} $\lambda$=64,000, ImageNet models: $\lambda$=150,000, Time-Series models: $\lambda$=32,000

    \item Parameter $\mathbf{W}$ for tiling and $\mathbf{A}$ for calculating $\alpha$.  $\mathbf{W}$ is used to learn a tile for each layer; it can also be used to calculate $\alpha$ scalars. Alternatively, we propose a separate parameter $\mathbf{A}$ to compute $\alpha$s independently, which exhibits a small performance gain. \textbf{Default:} $\mathbf{A}$ for calculating $\alpha$.  For ImageNet, we use $\mathbf{W}$.  

        \item Tile-wise $\alpha$s.  We experiment with calculating a single $\alpha$ per layer as well as calculating $\alpha$ for each tile in a layer. In some settings multiple $\alpha$s perform better.  \textbf{Default:} Single $\alpha$ per layer. 
\end{enumerate}

\section{Experiments}\label{experiments}
\begin{table}[t]
\scalebox{1.1}{
\centering\arraybackslash
\begin{tabular}{c|c c c c} 
\hline
  \multicolumn{5}{c}{\thead{\textbf{CIFAR-10}} }\\    \hline
 Model & Method & \thead{Bit-Width\\ (Savings)}  & \thead{\#Params\\ (M-Bit)} & \thead{Test Acc.\\ (\%)} \\
 \hline
{\multirow{9}{*}{\thead{ResNet\\18}}}
 &\footnotesize{Full-Precision} & 32 & 351.54 & 93.1\\
&IR-Net & 1 & 10.99 & 92.9\\
  &SNN & 0.440 \footnotesize{\textcolor{blue}{(2.3x)}}& 4.88 & 92.1\\
      &Sparks* &0.440 \footnotesize{\textcolor{blue}{(2.3x)}}&4.88& 90.8\\
  &MST* &0.075 \footnotesize{\textcolor{blue}{(13.3x)}}& 0.81& 91.6 \\
\cline{2-5}
&\cellcolor{aliceblue}\gls{tbn}\textsubscript{4}&\cellcolor{aliceblue}0.256 \footnotesize{\textcolor{blue}{(3.9x)}}&\cellcolor{aliceblue}2.85&\cellcolor{aliceblue}93.1 \\
&\cellcolor{aliceblue}\gls{tbn}\textsubscript{8}&\cellcolor{aliceblue}0.131 \footnotesize{\textcolor{blue}{(7.7x)}} &\cellcolor{aliceblue}1.46& \cellcolor{aliceblue}92.4\\
&\cellcolor{aliceblue}\gls{tbn}\textsubscript{16}&\cellcolor{aliceblue}0.069 \footnotesize{\textcolor{blue}{(14.5x)}}&\cellcolor{aliceblue}0.77 &\cellcolor{aliceblue} 91.2 \\
 \hline
{\multirow{6}{*}{\footnotesize{\thead{ResNet\\50}}}}
 &\footnotesize{Full-Precision} & 32 &  750.26&95.4\\
&IR-Net & 1 & 23.45&93.2\\
 &SNN & 0.35 \footnotesize{\textcolor{blue}{(2.8x)}}& 8.32& 94.0 \\
 \cline{2-5}
&\cellcolor{aliceblue}\gls{tbn}\textsubscript{4}&\cellcolor{aliceblue}0.259 \footnotesize{\textcolor{blue}{(3.9x)}} &\cellcolor{aliceblue}6.10&\cellcolor{aliceblue}94.9\\
&\cellcolor{aliceblue}\gls{tbn}\textsubscript{8}&\cellcolor{aliceblue}0.136 \footnotesize{\textcolor{blue}{(7.4x)}} &\cellcolor{aliceblue}3.21&\cellcolor{aliceblue}94.3 \\
&\cellcolor{aliceblue}\gls{tbn}\textsubscript{16}&\cellcolor{aliceblue}0.075 \footnotesize{\textcolor{blue}{(13.3x)}}&\cellcolor{aliceblue}1.76&\cellcolor{aliceblue}93.5 \\
\hline
 {\multirow{7}{*}{\footnotesize{\thead{VGG\\Small}}}}
 &\footnotesize{Full-Precision} & 32 &  146.24 & 92.7\\
&IR-Net & 1 & 4.656 & 91.3\\
&SNN& 0.440 \footnotesize{\textcolor{blue}{(2.3x)}}&2.032&91.9\\
&Spark*&0.440 \footnotesize{\textcolor{blue}{(2.3x)}}&2.032&90.8 \\
\cline{2-5}
&\cellcolor{aliceblue}\gls{tbn}\textsubscript{4}&\cellcolor{aliceblue} 0.288 \footnotesize{\textcolor{blue}{(3.5x)}} &\cellcolor{aliceblue} 1.340 &\cellcolor{aliceblue} 92.6\\
&\cellcolor{aliceblue}\gls{tbn}\textsubscript{8}&\cellcolor{aliceblue}0.131 \footnotesize{\textcolor{blue}{(7.7x)}}& \cellcolor{aliceblue}0.722&\cellcolor{aliceblue}91.5\\
&\cellcolor{aliceblue}\gls{tbn}\textsubscript{16}&\cellcolor{aliceblue}0.117 \footnotesize{\textcolor{blue}{(8.6x)}}&\cellcolor{aliceblue}0.520&\cellcolor{aliceblue} 90.2 \\
 \hline
  \multicolumn{5}{c}{\thead{\textbf{ImageNet}} }\\    \hline
  {\multirow{6}{*}{\footnotesize{\thead{ResNet\\34}}}}
&\footnotesize{Full-Precision}&32&674.88&73.1\\
&IR-Net&1&21.09&70.4\\
&SNN&0.560 \footnotesize{\textcolor{blue}{(1.8x)}}&11.71 &66.9\\
&MST*&0.450 \footnotesize{\textcolor{blue}{(2.2x)}}&9.51 &65.4\\
&Sparks*&0.560 \footnotesize{\textcolor{blue}{(1.9x)}}&11.71 &67.6 \\
\cline{2-5}
&\cellcolor{aliceblue}\gls{tbn}\textsubscript{2}& \cellcolor{aliceblue}0.53 \footnotesize{\textcolor{blue}{(1.9x)}}&\cellcolor{aliceblue}11.13 &\cellcolor{aliceblue}68.9 \\
\hline
\end{tabular}}
\caption{\textbf{CNN Results on CIFAR-10 and ImageNet}:  * indicates model with binary activations. We denote the tiling compression $p$ of each experiment as \gls{tbn}\textsubscript{p}. Savings (in \textcolor{blue}{blue}) indicates the compression from a binary-weight model (1-bit per model parameter). }\label{cifar10results}
\end{table}

We next detail our experiments across a range of architectures, datasets, and tasks.  We test \glspl{tbn} on CNNs as well as fully-connected models such as PointNet and Transformers. 

\vspace{-1em}

\subsection{CNN Architectures }

In this section,  we compare \glspl{tbn}  against previous sub-bit compression techniques for \glspl{cnn} including SNN \citep{wang2021sub}, MST \citep{vo2023mst}, and Spark \citep{wang2023compacting}. CNN's are the only models which have achieved sub-bit compression in previous research and thus they are a strong choice for benchmarking \glspl{tbn}. We note that \glspl{tbn} also work on Transformers and MLPs, which we experiment with in the next sections. 

We assess \glspl{tbn} on both CIFAR-10 and ImageNet datasets using the ResNet18/34 models \citep{he2016deep}, similar to previous works.  In addition, we include ResNet50 for CIFAR-10, which has  $1 \times 1$ convolutional kernels.  The SNN sub-bit compression algorithm was assessed with Resnet50 using a modified kernel selection technique specialized for  $1 \times 1$ convolutions, enabling up to 8x compression.  The technique, from Section C of the Appendix in \citep{wang2021sub}, is the most similar approach to TBNs. 

In Table \ref{cifar10results} we compare the performance of \glspl{tbn} to previous approaches using bit-width, number of parameters, and test set accuracy.  Bit-width measures the number of bits per model parameter, with a full precision model having 32-bits per parameter and a binary model having 1-bit per parameter. In blue, we include the savings of \glspl{tbn} compared to binary neural networks. In the next we column we denote the number of parameters required to save each model for inference, and finally the test set accuracy is the percentage of correct predictions by the model. Unique to the CNN architecture, we additionally include the number of bit operations required for each model in the results section.

\textbf{Results.}
Table \ref{cifar10results} highlights the results of \glspl{tbn} compared to previous sub-bit compression techniques on the CIFAR-10 and ImageNet datasets.  For CIFAR-10, we achieve sub-bit compression across ResNet architectures without a decrease in test performance at 4x compression. Experiments are run three times each and averaged.  Compared to other methods, \glspl{tbn} achieves a competitive performance with MST, the current state-of-the-art method for sub-bit compression.  \glspl{tbn} achieve similar performance at the same compression rates of MST for the CIFAR-10 models.  For ImageNet, \glspl{tbn} achieves enhanced performance roughly 2x compression, with test accuracy within 1.5\% of the performance of the binary-weighted IR-Net.  

In addition to the reduction in bit-width and parameter count, \glspl{tbn} trained with our default training configurations (single $\alpha$, no bias parameters) create replicated convolutional channels.  For example a 2d convolutional layer with one input channel, two output channels, $3 \times 3$ kernel size, and a tile compression rate of $p = 2$ will create two identical output channels. This means that during inference \glspl{tbn} lead to a substantial reduction in the number of required convolutional computations -- only one of the tile computations need to be executed, and we can replicate output channels from the other tiles.  Table \ref{bitops} summarizes the bit-ops saving achieved by \glspl{tbn} compared to both full precision on binary neural networks.  

\vspace{1em}

\begin{table}[h]
\centering
\scalebox{0.98}{
\begin{tabular}{c|c|c|c|c}
\hline
\textbf{Dataset} &\textbf{Dataset} & \textbf{Full Precision} & \textbf{IR-Net} & \textbf{\gls{tbn}} \\ \hline
{\multirow{2}{*}{\thead{CIFAR-10}}}& ResNet18  & 35.03                  &  0.547           & 0.082\cellcolor{aliceblue}  \footnotesize{\textcolor{blue}{(6.7x)}}  \\ 
& ResNet50  &  78.12                  &  1.22           & 0.155\cellcolor{aliceblue}     \footnotesize{\textcolor{blue}{(7.9x)}}    \\ \hline
\footnotesize{ImageNet}& ResNet34 & 225.66                & 3.526          & 0.58\cellcolor{aliceblue}     \footnotesize{\textcolor{blue}{(6.1x)}}    \\ \hline
\end{tabular}
}
\caption{Bit-Ops of ResNet architectures trained with a) full precision, b) IR-Net, and c) \gls{tbn}.  CIFAR-10 \glspl{tbn} have $p=4$, and ImageNet \glspl{tbn} has $p=2$. } \label{bitops}
\end{table}

\begin{table}[t]
\scalebox{1.1}{
\centering

\begin{tabular}{  c c c c c} 
 \hline
\hline
  \multicolumn{5}{c}{\thead{\textbf{Task: Classification, Dataset: ModelNet40}} }\\  
  \hline\hline
  
Algorithm &  \thead{Bit-Width\\ (Params)} & \thead{\#Params\\ (M-Bit)} &\multicolumn{2}{c}{\thead{Test Acc.\\ (\%)} }\\

 \hline
 

\footnotesize{Full-Precision}&32 &111.28 &\multicolumn{2}{c}{90.30} \\

FDA*& 1& 3.48 & \multicolumn{2}{c}{81.87}\\
\gls{bwnn}& 1&3.48& \multicolumn{2}{c}{89.20} \\
\cline{1-5}
\cellcolor{aliceblue}\gls{tbn}\textsubscript{4}&\cellcolor{aliceblue}0.259 \footnotesize{\textcolor{blue}{(3.9x)}} &\cellcolor{aliceblue}0.90&\multicolumn{2}{c}{\cellcolor{aliceblue}88.67}\\
\cellcolor{aliceblue}\gls{tbn}\textsubscript{8}&\cellcolor{aliceblue} 0.136 \footnotesize{\textcolor{blue}{(7.4x)}} &\cellcolor{aliceblue}0.47&\multicolumn{2}{c}{\cellcolor{aliceblue}87.20}\\
\hline\hline

  \multicolumn{5}{c}{\thead{\textbf{Task: Part Segmentation, Dataset: ShapeNet}} }\\    \hline\hline
  Algorithm & \thead{Bit-Width\\ (Params)} & \thead{\#Params\\ (M-Bit)} &\thead{Instance\\Avg. IoU}&\thead{Class\\Avg. IoU}\\
 \hline

\footnotesize{Full-Precision}&32 &266.96 &83.06&77.43 \\
XNOR-Net*&1 & 8.34& - & 60.87\\
\gls{bwnn}& 1& 8.34&76.1&69.90\\
\cline{1-5}
\cellcolor{aliceblue}\gls{tbn}\textsubscript{4}&\cellcolor{aliceblue}0.340 \footnotesize{\textcolor{blue}{(2.9x)}} &\cellcolor{aliceblue}2.68&\cellcolor{aliceblue}76.3&\cellcolor{aliceblue}70.20\\
\cellcolor{aliceblue}\gls{tbn}\textsubscript{8}&\cellcolor{aliceblue}0.207 \footnotesize{\textcolor{blue}{(4.8x)}} &\cellcolor{aliceblue}1.73&\cellcolor{aliceblue}75.1&\cellcolor{aliceblue}68.90\\

\hline
\hline
  \multicolumn{5}{c}{\thead{\textbf{Task: Semantic Segmentation, Dataset: S3DIS}} }\\    \hline\hline
  Algorithm & \thead{Bit-Width\\ (Params)} & \thead{\#Params\\ (M-Bit)} &\thead{Test\\Acc.}&\thead{Class\\Avg. IoU}\\
 \hline

\footnotesize{Full-Precision}& 32&112.96&78.27&42.20 \\
\glspl{bwnn}&1 &3.53& 69.50&31.30\\
\cline{1-5}
\cellcolor{aliceblue}\gls{tbn}\textsubscript{4}&\cellcolor{aliceblue} 0.431 \footnotesize{\textcolor{blue}{(2.3x)}} 
 &\cellcolor{aliceblue}1.52&\cellcolor{aliceblue}67.55&\cellcolor{aliceblue}31.10\\
\cellcolor{aliceblue}\gls{tbn}\textsubscript{8}&\cellcolor{aliceblue}0.337 \footnotesize{\textcolor{blue}{(3.0x)}}  &\cellcolor{aliceblue}1.19&\cellcolor{aliceblue}65.70&\cellcolor{aliceblue}29.55\\

 \hline

 \hline
\end{tabular}}
\caption{\textbf{PointNet Results}: We test \glspl{tbn} on the fully-connected PointNet model. \glspl{tbn} achieve performance close to the full-precision model on the classification benchmark and within 10\% of full-precision performance on the Part Segmentation task. * indicates model results from BiBench with binary activations. We take BiBench binarization algorithm with the best results for both ModelNet40 and ShapeNet.}
\label{pointnet_results}
\end{table}

\vspace{-2em}

\subsection{MLP-Based Architectures}

In addition to CNN architectures, we consider MLP models which contain a high proportion of fully-connected and $1 \times 1$ convolutional layers.  PointNet is a well-established model for unified tasks in classification, part segmentation, and semantic segmentation \citep{qi2017pointnet}. The model takes point cloud data from 3D representations.  To assess PointNet we use datasets ModelNet40, Shapenet, and S3DIS, which are each designed for a specific task. 
We denote segmentation performance with Intersection over Union (IoU), and class average IoU.  IoU is the ratio of the intersection area between the predicted and ground truth regions to the union area of both regions. Average IoU is calculated across all instances or regions in the dataset, whereas class average IoU is calculated separately for each class and then averaged across all classes.


We derive MLP experiments from BiBench \citep{10.5555/3618408.3619585}, who provide a diverse set of tasks to evaluate \glspl{bnn}. 
We note that the benchmarks provided in BiBench assess binarizing pretrained models.  Additionally BiBench only assesses models with binary weights and activations. 
In Table \ref{pointnet_results} we denote the best algorithms from BiBench. We additionally train a \gls{bwnn} for each task for a stronger comparison. 

\textbf{Results.} In Table \ref{pointnet_results} we summarize the PointNet model results across classification (ModelNet40), part segmentation (ShapeNet), and semantic segmentation (S3DIS).  We find that classification performance is almost on par with the full-precision model, with less than a 2\% drop in accuracy at 4x compression. For both part and semantic segmentation, \glspl{tbn} exhibit some loss in accuracy compared to their full-precision counterpart.  However, we note that in both cases \glspl{tbn} perform on par with \glspl{bwnn}. \glspl{tbn} also significantly outperform XNOR-Net in part segmentation, the most successful \gls{bnn} on the task in BiBench.

\begin{table}[t]
\centering
\scalebox{1.1}{
\begin{tabular}{c|c c c c} 
 \hline

  \multicolumn{5}{c}{\thead{\textbf{CIFAR-10}}} \\    
  \hline
 Model & Method & \thead{Bit-Width\\ (Params)} & \thead{\#Params\\ (M-Bit)} & \thead{Test Acc.\\ ($\%$)} \\ [0.5ex] 
 \hline
\multirow{4}{*}{\footnotesize{\thead{ViT}}}
 & \footnotesize{Full-Precision} & 32 & 303.68 & 82.5 \\
 & \footnotesize{\gls{bwnn}} & 1 & 9.50 & 82.2 \\
 \cline{2-5}
 & \cellcolor{aliceblue}\gls{tbn}\textsubscript{4} & \cellcolor{aliceblue} 0.253 \footnotesize{\textcolor{blue}{(4.0x)}} & \cellcolor{aliceblue} 2.40 &\cellcolor{aliceblue} 82.7 \\
 & \cellcolor{aliceblue} \gls{tbn}\textsubscript{8} & \cellcolor{aliceblue}0.129 \footnotesize{\textcolor{blue}{(7.8x)}} &\cellcolor{aliceblue} 1.22 &\cellcolor{aliceblue} 82.1 \\
 \hline
\multirow{4}{*}{\footnotesize{\thead{Swin-t}}}
 & \footnotesize{Full-Precision} & 32 & 851.14 & 86.8 \\
 & \footnotesize{\gls{bwnn}} & 1 & 26.60 & 85.8 \\
 \cline{2-5}
 & \cellcolor{aliceblue}\gls{tbn}\textsubscript{4} & \cellcolor{aliceblue}0.259 \footnotesize{\textcolor{blue}{(3.9x)}} &\cellcolor{aliceblue} 6.88 & \cellcolor{aliceblue}85.8 \\
 & \cellcolor{aliceblue} \gls{tbn}\textsubscript{8} & \cellcolor{aliceblue}0.135 \footnotesize{\textcolor{blue}{(7.4x)}} & \cellcolor{aliceblue}3.61 & \cellcolor{aliceblue}84.6 \\
 \hline

  \multicolumn{5}{c}{\thead{\textbf{ImageNet}}} \\    
  \hline\hline
\multirow{2}{*}{\footnotesize{\thead{Swin-t}}}
 & \footnotesize{Full-Precision} & 32 & 873.60 & 81.3 \\
 \cline{2-5}
 & \cellcolor{aliceblue} \gls{tbn}\textsubscript{2} &\cellcolor{aliceblue} 0.534 \footnotesize{\textcolor{blue}{(1.9x)}} &\cellcolor{aliceblue} 14.7 &\cellcolor{aliceblue} 77.3 \\
 \hline
\end{tabular}
}
\caption{\textbf{Vision Transformers trained on CIFAR-10 and ImageNet}: We compare the performance of a \gls{vit} (patch size 4) and the Swin-t model with \gls{tbn}\textsubscript{4} and \gls{tbn}\textsubscript{8} variations for CIFAR-10 and \gls{tbn}\textsubscript{2} for ImageNet. \textsuperscript{$\dagger$} The Swin-t \gls{tbn}\textsubscript{2} ImageNet model was trained for 200 epochs with basic data augmentations. }\label{transformers}
\end{table}


\subsection{Transformers}\label{vision_transformers}
In our next set of experiments, we assess \glspl{tbn} on Transformers.  It was noted in the \gls{bnn} benchmark work BiBench \citep{10.5555/3618408.3619585} that Transformers perform poorly on \glspl{bnn}. In particular, none of the binarization algorithms tested in BiBench achieved more than 70\% of the performance of full-precision models.  The research, which looks exclusively at models with binary weights \textit{and} activations, noted that binarization of activations greatly affects the attention mechanism and leads to poor quality models.  The equations between the query key, and value cause amplified information loss when model activations are binarized. 
In contrast to BiBench, we assess models with binary weights and full-precision activations which do not suffer from the same information loss as models with binary activations.

We test Transformer \glspl{tbn} on three tasks: 1) Vision Transformers, 2) Small NLP Transformers (encoders), and 3) time-series Transformers (encoders). 

\subsubsection{Vision Transformers}
For Vision Transformers
 We train a \gls{vit} \citep{dosovitskiy2020image}, which uses image patching along with the traditional attention, and Swin-t \citep{liu2021swin}, which uses hierarchical partitioning of the image into patches with merging.  We train the models from scratch on the CIFAR-10 and ImageNet datasets. Note that, training the models from scratch, rather than fine-tuning, can result in sub-optimal performance in Transformers \citep{dosovitskiy2020image}.  

\textbf{Results.} 
In Table \ref{transformers} we summarize the results of \glspl{vit} trained from scratch. \glspl{tbn} achieve performance within 2.5\% of full-precision models across all CIFAR-10 experiments. Specifically, the \gls{tbn} \gls{vit} can closely match the accuracy of the full-precision model at both 4x and 8x compression rates, while Swin-t has a performance degradation of just 1.2\% at 4x and 2.5\% at 8x. For ImageNet, the Swin-t  \gls{tbn} quickly converged to a stronger accuracy than the ResNet34 ImageNet model in just 200 epochs.  Moreover, we only trained the model with basic data augmentations. In contrast, the full-precision Swin-t model was trained for 300 epochs with enhanced data augmentation techniques.

\subsubsection{Time Series Transformers}
In our next set of experiments we explore Transformer \glspl{tbn} for multivariate time series forecasting.  Transformer encoders have been shown to be robust to model compression for time series learning tasks including forecasting \citep{gorbett2023sparse}; we aim to assess whether we can compress similar models using \glspl{tbn}.

Our experiments utilize the ECL (Electricity) dataset and the Weather dataset similar to previous works \citep{gorbett2023sparse, zhou2021informer}.  ECL has 321 features and Weather has 7 features. We use a Transformer encoder similar to previous work \citep{zerveas2021transformer}, using a dimensionality of 512 for the Electricity dataset and 128 for the Weather dataset. Similar to previous work, we assess the performance of the models use Mean Squared Error.  

\textbf{Results.}  We train each of the models five times and report the average MSE along with the standard deviation in Table \ref{time}. Notably, \glspl{tbn} performed slightly better than full precision and binary models on the electricity dataset.  On the weather dataset, we set $\lambda$=32,000 since all the layers are smaller than this. We find that the model is able to converge to full precision accuracy.  However, when we lower $\lambda$ to 16,000 model performance suffers.

\begin{table}[h]
\centering
\begin{tabular}{c|cccc}
\hline
 Dataset & Method & \thead{Bit-Width\\ (Params)} & \thead{\#Params\\ (M-Bit)} & MSE\\  
 \hline
{\multirow{3}{*}{\footnotesize{\thead{Electricity}}}}
&\footnotesize{Full-Precision}&32&145.2& 0.212$\pm$0.10 \\
& \footnotesize{\gls{bwnn}} &1&4.5 & 0.210$\pm$0.05\\
\cline{2-5}
&\cellcolor{aliceblue}\gls{tbn}\textsubscript{4}  &\cellcolor{aliceblue}0.25 \footnotesize{\textcolor{blue}{(4.0x)}}&\cellcolor{aliceblue}1.1 &\cellcolor{aliceblue} 0.209$\pm$0.13 \\
\hline
{\multirow{3}{*}{\footnotesize{\thead{Weather}}}}
&\footnotesize{Full-Precision}&32&11.8& 0.165$\pm$0.10 \\
& \footnotesize{\gls{bwnn}} &1&0.368 & 0.165$\pm$0.02\\
\cline{2-5}
&\cellcolor{aliceblue}\gls{tbn}\textsubscript{4} &\cellcolor{aliceblue}0.54 \footnotesize{\textcolor{blue}{(1.9x)}} &\cellcolor{aliceblue}0.197 & \cellcolor{aliceblue}0.168$\pm$0.12\\
\hline
\end{tabular}
\caption{Multivariate Time Series Forecasting: We compare \glspl{tbn} to full precision neural networks and binary weighted neural networks (\glspl{bwnn}) on two multivariate time series datasets. We find that \glspl{tbn} achieve similar performance to larger models on single-step forecasting tasks.   } \label{time}
\end{table}

\section{Implementation}\label{imp}

We cover two implementations of \glspl{tbn} for model inference: 1) a C implementation which we deploy on a microcontroller, and 2) GPU-compatible inference kernels. Both methods implement \textit{reuse} of a single tile per layer to achieve memory and storage savings. 

\subsection{Microcontroller Deployment }

We first implement an inference engine to run on an Arduino microcontroller.  The microcontroller has 1MB of storage and 250KB of memory, making it practical for a lightweight model. We program an MLP model trained on the MNIST dataset with 128 hidden neurons and a fused ReLU nonlinearity. We implement both a standard \gls{bwnn} and a \gls{tbn}. 
Our \gls{tbn} has a compression rate of 4 and uses multiple $\alpha$s (one for each tile). 
To implement the \gls{tbn} we first train our model in PyTorch \citep{paszke2019pytorch}, and then convert the layer tiles and $\alpha$ scalars to C data types.  We implement a fully-connected kernel in C as detailed in Algorithm \ref{fc_kernel}. We develop a fully binarized kernel by packing binary weights into unsigned 8-bit integers and use bit-masking to extract the correct values during inference.

We assess the speed, memory, and storage space of each model.  To assess speed, we report the \gls{fps}, which measures the number of times the program can execute on a sample per second. We measure FPS using a provided script which executes the compiled model 1000 times and reports the mean and standard deviation across five runs. We report memory as the maximum memory at any layer of the model. Finally, we report the storage space as the number of bits stored for each model.  We expect speed to be the same across models, while memory and storage space should improve with \glspl{tbn}. 

\textbf{Results.} We summarize the results of the implementation across speed, memory, and storage space in Table \ref{perf}. As expected, the speed of both models is roughly the same. The max memory usage, which happens during execution of the first fully-connected layer, is 58\% less for the \glspl{tbn} compared to the \gls{bwnn}. Both models still require full-precision inputs and a full precision placeholder for the output, while the \glspl{tbn} model requires roughly $\frac{1}{4}$ of the weights loaded in memory compared to the \gls{bwnn}.  Finally, storage for the \gls{tbn} model is almost $\frac{1}{4}$ that of the \gls{bwnn}.  Since the classification layer only contains 1280 parameters, it is not tiled, accounting for more parameters in the tiled model.  

\begin{algorithm}[t]
\caption{FC Layer with Tiling, Many $\alpha$s, Forward Pass}
\begin{algorithmic}
\State \textbf{Inputs:} Tile vector $\mathbf{t}$ of size $q$, input vector $\mathbf{x}$ of size $n$, layer size metadata ($m, n$), $\alpha_1, \alpha_2, ... \alpha_p$
\State $t_i\gets 0, \quad \alpha_i\gets \alpha_1$ 
\For{$i \gets 1$ to $m$}
    \State $\mathbf{y}[i] \gets 0$ 
    \State $\alpha_i=\alpha_i+1$
    \For{$j \gets 1$ to $n$}
      \State $\mathbf{y}[i] \gets \mathbf{y}[i] + \mathbf{t}[{t_i}] \cdot \mathbf{x}[j]\cdot \alpha_i$ 

\If{$t_i = q$} 
    \State $t_i\gets0$ \Comment{ Move to beginning of tile vector}
    \State $\alpha_i\gets \alpha_{i+1}$  \Comment{ Get next tiles $\alpha$}
\Else
        \State $t_i \gets t_i+1$ \Comment{ Increment tile index}
\EndIf
    \EndFor
    \State $\mathbf{y}[i] \gets \max(0, \mathbf{y}[i])$   \Comment{ Fused ReLU}
  \EndFor
  \State \textbf{Output:} Output vector $\mathbf{y}$ of size $m$
\end{algorithmic}\label{fc_kernel}
\end{algorithm}

\begin{table}[h]
\centering
\renewcommand{\arraystretch}{1.1}
\begin{tabular}{c c c c} 
 \hline
Model & \thead{Speed\\(FPS)}  & \thead{Max Memory\\Usage (KB)} & \thead{Storage\\(KB)} \\ \hline
 \gls{bwnn}&704.5$\pm$3.3&16.20&12.70 \\
  \cline{1-4}
\rowcolor{aliceblue}\gls{tbn}\textsubscript{4}&705.1$\pm3.6$&6.80&3.32\\

 \hline
\end{tabular}
\caption{
We compare the performance of a binary-weight neural network (\gls{bwnn}) and a \gls{tbn} deployed on a microcontroller. We implement an MLP with one hidden layer (size 128). The maximum memory usage corresponds to the full-precision image being processed by the first fully-connected layer, with additional memory allocated for the output activation's.    } \label{perf}
\end{table}

\subsection{GPU Inference Kernel}

Our second implementation utilizes the open-source Triton library, which enables us to write customized CUDA kernels in Python to run on GPUs. Native PyTorch does not allow the reuse of a single tile without allocating new memory for the full layers parameters ($p$ tiles) -- memory reuse between tensors with incompatible strides is not possible in standard operations. Leveraging Triton provides more control over lower-level memory allocation. 

We implement fully-connected kernels using both full-precision (32-bit) weights as well as binary (1-bit) weights. For the former use case, we experiment tiling layers with full-precision weights to compare against standard 32-bit models. For the latter, we pack bit parameters into unsigned 8-bit integers, and unpack them during inference. Bit values are packed by \textit{row}, converting an $m \times n$ matrix to size $(\frac{m}{8}) \times n$. 

We implement both kernels using the matrix multiplication functionality provided by Triton. It uses block-level matrix multiplication and has a similar performance to the optimized cuBLAS library. Our tiling implementation converts an $m\times n$ matrix to $m\times q$ (we compress the second dimension). For pointer arithmetic, we reuse the same $m\times q$ tile for multiple computations. In other words, fully-connected tiling is a \textit{matrix-to-matrix} implementation rather than \textit{matrix-to-vector} implementation.


\begin{figure}[t]
\centering
\includegraphics[width=8cm]{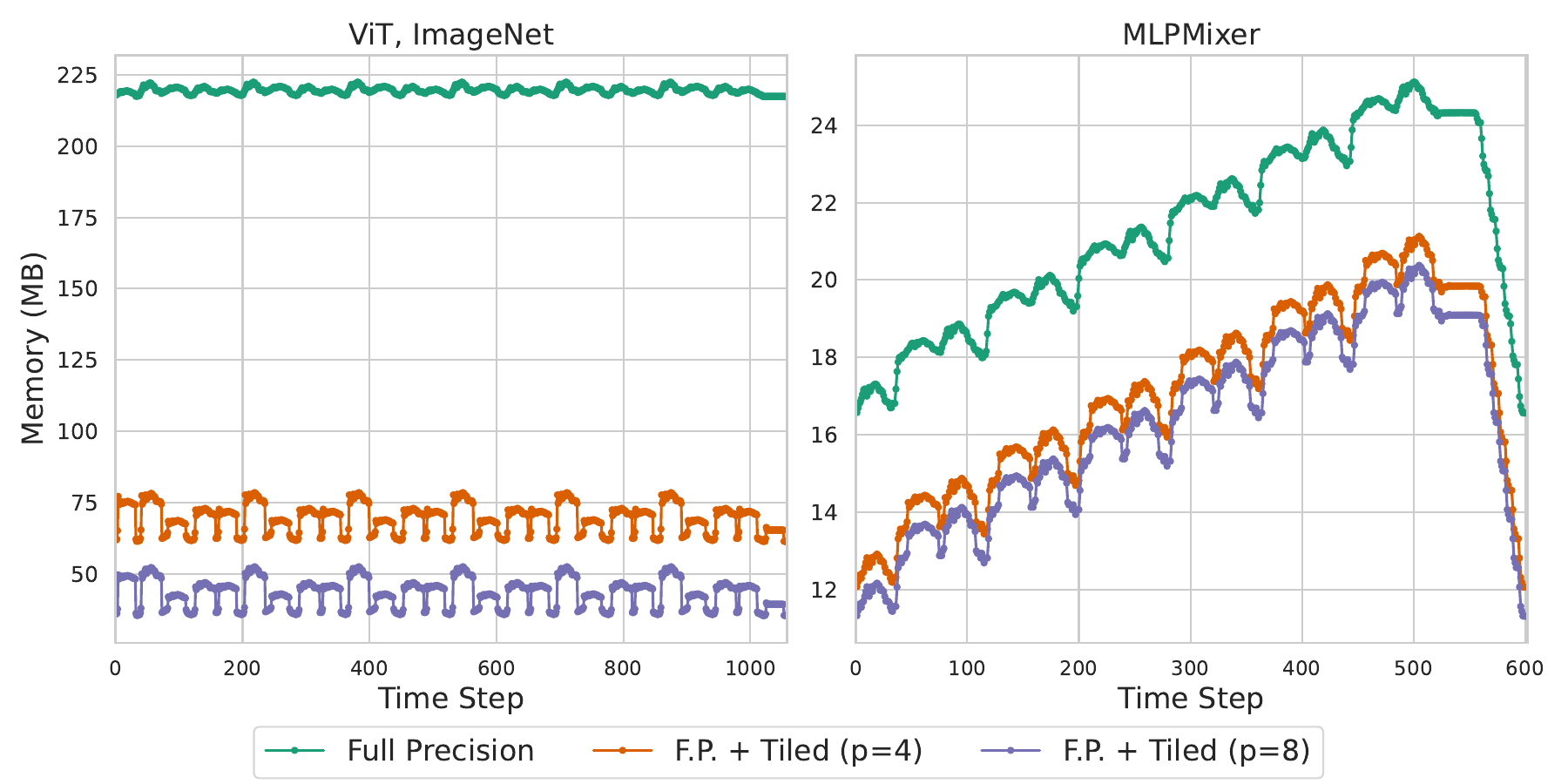}
\caption{ 
\textbf{GPU memory allocated during model inference:} We profile the memory of the ImageNet \gls{vit} (left) and PointNet (right) during inference using a customized GPU kernel with full-precision weights.  The x-axis represents memory recorded within intermediate model layers during execution of a PyTorch model. The tiled kernel achieves 2.8x memory reduction on the \gls{vit} and 1.2x reduction on PointNet. 
}\label{vit_mlpmixer}
\end{figure}

\begin{table}[b]
\scalebox{1}{
\centering
\renewcommand{\arraystretch}{1.1}
\begin{tabular}{c l l c} 
 \hline
Model & \thead{Peak Memory\\(MB)} & \thead{Parameter\\Mem. (MB)} & \thead{\% Param.\\Mem. (MB)} \\ \hline
\small{Full Precision}& 222.5&208.0&93.5\%\\
  \cline{1-4}
\rowcolor{aliceblue}\small{FP, Tiled$_4$}& 78.5 \footnotesize{\textcolor{blue}{(2.8x)}}&52.0 \footnotesize{\textcolor{blue}{(4.0x)}} &66.2\%\\
\hline
   \gls{bwnn}&18.4&6.5&35.3\%\\
     \cline{1-4}
\rowcolor{aliceblue}\gls{tbn}\textsubscript{4}&13.4 \footnotesize{\textcolor{blue}{(1.4x)}} & 1.6 \footnotesize{\textcolor{blue}{(4.1x)}} &11.9\%\\
 
 \hline
\end{tabular}
}
\caption{\textbf{Memory assessment during inference on ImageNet \gls{vit}:} Tiled kernels reduce the memory occupied by model weights, leading to a lower overall memory footprint. The final column is the \% of the peak memory used by weights, which is substantially lower for \glspl{tbn}. 
  } \label{gpu_perf}
\end{table}

\textbf{Results.} We measure the GPU memory usage for inference on a single image in Figure \ref{vit_mlpmixer} and Table \ref{gpu_perf}. Figure \ref{vit_mlpmixer} profiles GPU memory usage through each layer of the model, while Table \ref{gpu_perf} examines the peak memory and memory occupied by model parameters in an ImageNet \gls{vit}. 

The results in Figure \ref{vit_mlpmixer} reflect the memory savings by only loading a \textit{single} tile per model layer for the standard full-precision kernel as well as the tiled full-precision kernel. Specifically, during inference the PyTorch library loads weights for \textit{all} layers into memory. It allocates new memory for the input and output activations of each layer, and deallocates the activations when they are no longer needed. For example, \gls{vit} has six attention layers with roughly 8.4 million parameters each ($Q,K,V$ and the feed-forward parameters) for a total of 54.6 million parameters. The input and output activations to the attention layers have a size of 65,536. As a result, we can see the small spikes in the graph, which represent the allocation and de-allocation of activations. However, most of the memory is a result of the weights (34MB per attention layer).  

Table \ref{gpu_perf} examines the memory savings of \glspl{tbn} on both peak memory and memory allocated by model weights for the ImageNet \gls{vit}. We observe a 2.8x reduction in peak memory  for the full-precision kernel and 1.4x reduction in peak memory for a binary kernel. When assessing the memory occupied specifically by model weights, we observe that \glspl{tbn} reduce memory by 4x, and as a result require a substantially smaller proportion of peak memory compared to their standard counterparts: in the binary weight setting, \gls{tbn} weights occupy just 11.9\% of the total peak memory.

\section{Ablation Studies}\label{ablation}

\begin{figure}[t]
\centering
\includegraphics[width=7.5cm]{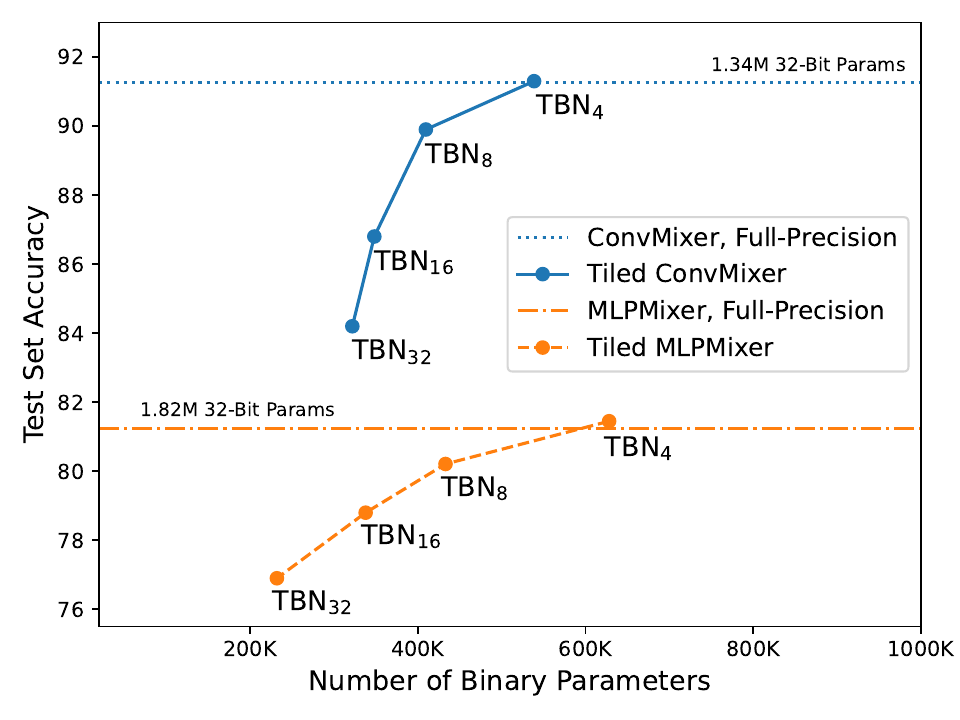}
\caption{ 
\textbf{\glspl{tbn} are vulnerable in models with low width layers:} We plot the performance of the ConvMixer and MLPMixer at various compression rates/number of parameters compared to its full-precision baseline (horizontal line).  The ConvMixer accuracy degrades quickly as a result of smaller layers: its maximum layer size is 65k.  The MLPMixer has layer sizes of 131k. Both models achieve near full-precision performance at 4x compression, and degrade at varying rates thereafter. 
}\label{teaser2}
\end{figure}


We perform several ablation study's on \glspl{tbn}, first exploring the effect of layer size on MLPMixers \citep{tolstikhin2021mlp} and ConvMixers \citep{trockman2022patches}, and then analyzing the effects of various hyperparameters in MLPMixers and ResNets.  

\subsection{Effects of Layer Size}

We next look at MLPMixers \citep{tolstikhin2021mlp} and ConvMixers \citep{trockman2022patches} for the CIFAR-10 classification task.  ConvMixer has shown potential to compete with \glspl{vit} on complex tasks, while MLPMixers provide us with another opportunity to test fully-connected models.  

In Figure \ref{teaser2} we visualize the performance of \glspl{tbn} at tiling compression rates up to 32x for both models. We find that the ConvMixer accuracy degrades quickly after 4x compression. When analyzing the architecture, we find that the largest layer has just 65,536 parameters. Moreover, many layers have less than 65k parameters and don't fulfill the minimum layer size for compression ($\lambda$), resulting in limited reductions in parameter count along with high performance degradation.  MLPMixer, on the other hand, has layers with 131k elements, and resulted in a more modest performance degradation at higher compression. Assessing other architectures, we found that the CNN, PointNet, and Transformer models contained layers significantly above $\lambda$. 

.

\begin{figure}[t]
\centering
\includegraphics[width=6.7cm]{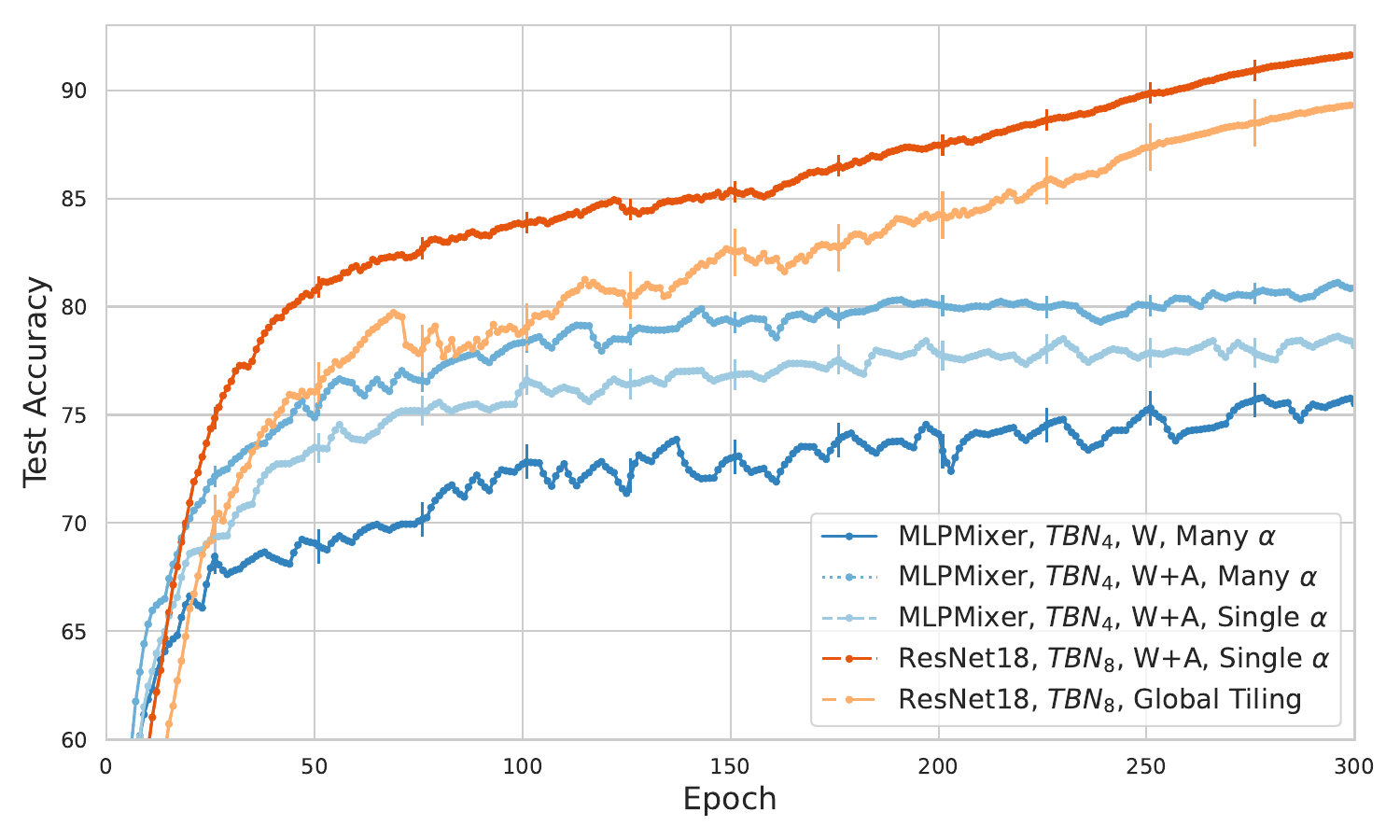}
\caption{\textbf{Hyperparameter Configurations:} We perform an ablation study to the assess the test set accuracy across training for the MLPMixer and ResNet18 models with various hyperparameter configurations.  For ResNet18, we show how tiling every convolutional layer, rather than layers of a certain size, leads to performance loss (red/orange). In blue we show the effects of using a separate parameter $\mathbf{A}$ to calculate $\alpha$ compared to calculating $\alpha$ using just $\mathbf{W}$. We additionally show the benefit of using multiple $\alpha$s (one per tile) rather than a single $\alpha$.
}\label{ablation2}
\end{figure}

\vspace{-1em}

\subsection{Hyperparameter Configuration Analysis}

We study the effects of three hyperparameters of \glspl{tbn} as described in Section \ref{methods}: 1) $\lambda$ (minimum layer size for tiling), 2) $\mathbf{W}$ for tiling and $\mathbf{A}$ for computing $\alpha$, and 3) Multiple versus single $\alpha$ scalars. 
For $\lambda$, we test global tiling, where all layers are compressed.  We compare this to our default training which sets $\lambda$ to 64k.  For the second hyperparameter, we test two settings: the first uses $\mathbf{W}$ for both learning the binary tile and calculating $\alpha$; the second uses parameter $\mathbf{A}$ for calculating $\alpha$.  We denote this setting as $\mathbf{W}$ or $\mathbf{W}+\mathbf{A}$. For the third parameter, we test a single $\alpha$ per layer and compare it to computing $\alpha$ for each tile.  We perform these experiments using the $\mathbf{W}+\mathbf{A}$ setting.  

Figure \ref{ablation2} shows the effects of the different hyperparameters on both the ResNet50 and MLPMixer models on the test accuracy throughout training.  We show that a global tiling factor on ResNet50 causes a significant decrease in performance. Next, we find the model converges best when given a separate parameter $\mathbf{A}$ to calculate $\alpha$.  Moreover, when calculating an $\alpha$ for each tile, we observe a slight increase in performance. 




\section{Discussion}
We propose \glsdisp{tbn}{Tiled Bit Networks} for learning binary vectors to fill in the weights of a neural networks layers during training, achieving sub-bit compression on model parameters during inference. \glspl{tbn} work on both convolutional and fully-connected layers, and are applicable to CNNs, MLPs, and Transformer-based architectures. 

Tiled Bit Networks have the potential to democratize deep learning by providing resource constrained devices access to larger scale models. For example, \glspl{tbn} compress the 23.5 million parameter ResNet50 model to under 6.1 million bits, a size which can fit into a microcontroller with 1MB of storage. Moreover, the algorithm unlocks potential for stronger sub-bit compression on other architectures, particularly those with fully-connected layers. The innovation not only increases the accessibility of deep learning use but also has the potential to contribute to environmental sustainability by reducing the carbon footprint associated with model size and computational complexity.

\textbf{Future Work}
A natural direction for future work is to apply \glspl{tbn} to models with binary weights and binary activations.  \glspl{bnn} with binary activations maximize memory savings and improve speed. We would also like to test the approach on larger models (e.g. LLMs), where additional algorithmic modifications (such as full-precision parameter tiling) may enhance performance. Additionally, specialized kernels could be implemented to maximize the efficiency of \glspl{tbn} with regards to parallelization, including convolutional kernels and kernels better optimized for the reusability of tiles.
In addition to direct algorithmic enhancements, \glspl{tbn} could also be assessed in parallel fields of machine learning such as adversarial detection \citep{goodfellow2014explaining, gorbett2022utilizing}, dataset complexity and \citep{gorbett2022wip,gorbett2022local, gorbett2023intrinsic}, and federated learning \citep{mcmahan2017communication,gorbett2023cross}.  







\setcounter{section}{0}
\section*{Appendix}

\renewcommand\thesection{\Alph{section}}
\renewcommand\thesubsection{\thesection.\Alph{subsection}}

\section{Training Details}
In this section we summarize the training details of our various experiments. 

\textbf{CNN's trained with CIFAR-10}
ResNet18, ResNet50, and VGG Small models are trained with similar hyperparameters on the CIFAR-10 dataset.  They use an SGD optimizer with a base learning rate of 0.1, weight decay 0.0001, and momentum 0.9. Batch size is 128. We use a cosine learning rate policy. Layers are created with kaiming normal initialization with scale fan.  We additionally use standard batch normalization. We normalize the dataset with mean and standard deviation. 
We initialize $\mathbf{W}$ and $\mathbf{A}$ with different seeds for each run (incremented by one for each experiment). 
\gls{tbn} hyperparameters: We use multiple $\alpha$'s by default, $\mathbf{W}$+$\mathbf{A}$. Our CIFAR-10 and ImageNet models are based off of the  Edge-Popup repository. 

\textbf{CNN's trained with ImageNet}
We train ResNet models on the ImageNet dataset using \glspl{tbn}.  We use baselines from other papers and literature.  We attain our default hyperparameters from Edge-Popup \citep{9157673}. We train our model for 600 epochs, using an SGD optimizer, learning rate of 0.256, a weight decay of 0.000030517578125, momentum of 0.875, and batch size 128.  We use a 0.1 value for label smoothing.  We additionally use a warmup length of 5 to keep the learning rate static while training starts. 

\textbf{PointNet Classification}
For our three PointNet tasks, we use an existing GitHub repository which you can find  \href{https://github.com/yanx27/Pointnet_Pointnet2_pytorch/tree/master}{here}.
We use a batch size of 24, and train the model for 500 epochs.  An Adam optimizer is used with a learning rate of 0.001.  We use a decay rate of 0.0001. 

\textbf{PointNet Part Segmentation}
We train our model for 250 epochs using a batch size of 16.  We use an Adam optimizer with a learning rate of 0.001 and a decay rate of 0.0001. We adjust the learning rate by 0.5 every 20 steps.  We use 2048 points (a hyperparameter for the ShapeNet dataset). 

\textbf{PointNet Semantic Segmentation}
We train our model for 32 epochs and a batch size of 16. We use an Adam optimizer with a learning rate 0.001, and a weight decay of 0.0001. We use 4096 points (a hyperparameter for the S3DIS dataset). We decay the learning rate by 0.7 every 10 steps. We use a test area of 5.  

\textbf{Transformers and Mixer Models trained with CIFAR-10}
We train our models between 500 to 1000 epochs. We use an Adam optimizer with a learning rate of 0.001. We use a batch size 128
For data augmentation, we use random cropping, random horizontal flip, and standard normalization for CIFAR-10. MLPMixer has a depth of 6, 512 output size, and a patch size of 4. Our Vision Transformer uses a patch size of 4,  a dimension head of 512, 8 heads, an MLP head of 512, and an embedding dropout of 0.1. 
For ConvMixer we use a convolutional kernel of 8, a patch size of 1, dimension of 256, and a depth of 16. For Swin-t we use downscaling factor of (2,2,2,1). 

\textbf{Swin-t with ImageNet}
We mostly follow the same training protocol as the original paper except with more epochs. We train for 600 epochs, using an AdamW optimizer, a cosine decay learning rate
scheduler, and 20 epochs of linear warm-up. We use a batch
size of 350, an initial learning rate of 0.001, and a
weight decay of 0.05. We use the same data augmentations as are used in the ResNet models.  



\section{Additional Ablation Study}

\begin{figure}[b]
\centering
\includegraphics[width=7cm]{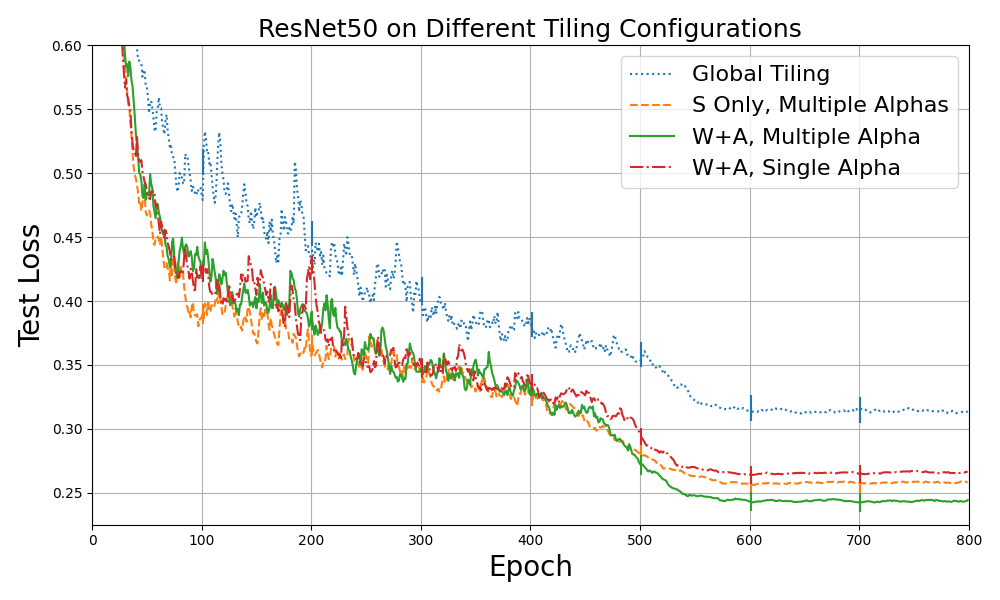}
\caption{ Test Loss on various ResNet-50 tiling configurations with compression rate 4. Using a minimum layer size for compression, multiple $\alpha$s, and a separate parameter $\mathbf{A}$ for calculating $\alpha$s performed the best (green line).  However, the only clear performance benefit across the hyperparameters came from using a minimum layer size for tiling $\lambda$. 
}\label{ablation3}
\end{figure}

We include an additional ablation in Figure \ref{ablation3}. Specifically, we look at the ResNet50 model with different tiling configurations such as global tiling versus minimum layer size tiling, $\mathbf{W}$ for calculating $\alpha$ versus $\mathbf{W}$ for tiling and $\mathbf{W}$ for calculating $\mathbf{A}$, and multiple versus one $\alpha$ parameter.  
We find that the configuration which clearly performs worst is global tiling.  As a result, we only tile layers which have a total size of at least 64,000.  For the other parameters, we notice a slight benefit when using multiple $\alpha$s (green line) compared to a single $\alpha$ (red).  The configuration which uses a separate parameter ($\mathbf{A}$) for calculating $\alpha$ performs better worse than the method which only uses $\mathbf{W}$. 

\bibliographystyle{ACM-Reference-Format}
\balance
\bibliography{egbib}





\end{document}